\documentclass{article}

\usepackage{arxiv}

\usepackage[utf8]{inputenc} 
\usepackage[T1]{fontenc}    
\usepackage{hyperref}       
\usepackage{url}            
\usepackage{booktabs}       
\usepackage{amsfonts}       
\usepackage{nicefrac}       
\usepackage{microtype}      
\usepackage{lipsum}
\usepackage{graphicx}
\graphicspath{ {./images/} }
\usepackage{hyperref}     
\usepackage{cleveref}     
\usepackage{setspace}
\usepackage{subcaption}
\usepackage{caption} 
\usepackage{graphicx}
\usepackage{wrapfig}

\title{Early Prediction of Sepsis: Feature-Aligned Transfer Learning}

\author{
 Oyindolapo O. Komolafe \\
  School of Physical Therapy,\\
  Faculty of Health Sciences,\\
  Western University, London, ON, CA \\
  \texttt{okomola@uwo.ca} \\
   \And
 Zhimin Mei \\
  Department of Computer Science,\\
  Western University\\
  London, ON, CA\\
  \texttt{zmei22@uwo.ca} \\
  \And
 David Morales Zarate \\
  Health and Rehabilitation Sciences\\
  Faculty of Health Sciences,\\
  Western University, London, ON, CA \\
  \texttt{omorales@uwo.ca} \\
   \And
 Gregory William Spangenberg \\
  Mechanical Engineering\\
   Western University\\
  London, ON, CA\\
  \texttt{gspangen@uwo.ca} \\
}
\begin{document}
\maketitle
\begin{abstract}
Sepsis is a life-threatening medical condition that occurs when the body has an extreme response to infection, leading to widespread inflammation, organ failure, and potentially death. Because sepsis can worsen rapidly—sometimes within hours—early detection is critical to saving lives. However, current diagnostic methods often identify sepsis only after significant damage has already occurred. Our project aims to address this challenge by developing a machine learning-based system to predict sepsis in its early stages, giving healthcare providers more time to intervene.

A major problem with existing models is the wide variability in the patient information—or features—they use, such as heart rate, temperature, and lab results. This inconsistency makes models difficult to compare and limits their ability to work across different hospitals and settings. To solve this, we propose a method called \textbf{Feature-Aligned Transfer Learning (FATL)}, which identifies and focuses on the most important and commonly reported features across multiple studies, ensuring the model remains consistent and clinically relevant.

Moreover, most existing models are trained on narrow patient groups, leading to population bias. FATL addresses this by combining knowledge from models trained on diverse populations, using a weighted approach that reflects each model’s contribution. This makes the system more generalizable and effective across different patient demographics and clinical environments. FATL offers a practical and scalable solution for early sepsis detection, particularly in hospitals with limited resources, and has the potential to improve patient outcomes, reduce healthcare costs, and support more equitable healthcare delivery. 
\end{abstract}

\keywords{sepsis \and  early prediction \and weight transfer \and transfer learning}

\section{Introduction}
\begin{figure}
    \centering
    \includegraphics[width=0.5\linewidth]{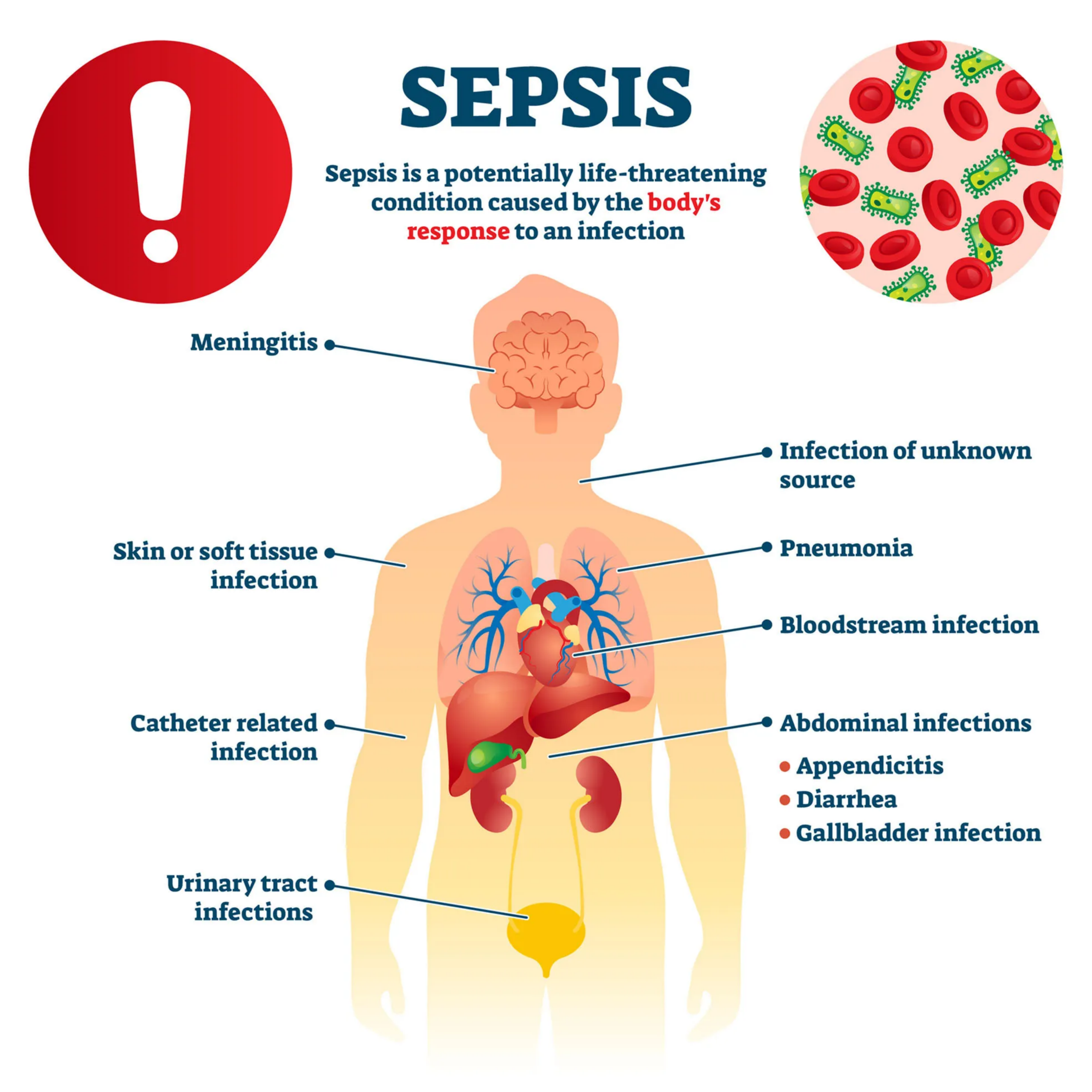}
    \caption{Introduction of sepsis}
    \label{fig:sepsis}
\end{figure}

Sepsis is a life-threatening condition that occurs when the body's immune system overreacts to an infection~\cite{Crighton2022}. Under normal circumstances, the body releases specific chemicals and proteins to regulate immune responses, control inflammation, and promote healing. However, in sepsis, this process becomes dysregulated, triggering an excessive and widespread inflammatory response throughout the body. Figure~\ref{fig:sepsis} shows the damage that sepsis can cause to different organs and tissues in the human body, reflecting its dangerousness.

The health risks associated with sepsis are severe and far-reaching. Often referred to as blood poisoning, sepsis can disrupt blood flow to vital organs and cause abnormal clotting, which may impair organ function. While early detection and treatment can often lead to recovery, untreated or advanced sepsis can progress rapidly to irreversible multiple organ failure, septic shock—the most critical phase—and ultimately, death.
\subsection{Societal and Clinical Importance of Sepsis}
Sepsis is a leading cause of critical illness and death worldwide~\cite{fleischmann2023challenges}. Estimates indicate 48.9 million cases of sepsis occurred in 2017, with 11 million or 20\% of deaths worldwide~\cite{rudd2020global}. The World Health Organization (WHO) has recognized sepsis as a global health priority and urged member states to improve its prevention, diagnosis, and management. Not only is sepsis deadly, but it is also costly. Managing sepsis often requires intensive care, prolonged hospital stays, and complex interventions. In high-income countries, the average hospital cost per sepsis case can exceed \$32,000 (USD)~\cite{arefian2017hospital}. According to Sepsis Canada, sepsis imposes an estimated \$1.7 billion (CAD) annually in direct costs~\cite{SepsisCanada2024}, with around \$1 billion per year in Ontario~\cite{farrah2021sepsis}. Beyond acute care, many sepsis survivors experience long-term consequences, which create additional societal costs for rehabilitation and support.

\subsection{Importance of early prediction of sepsis}
\begin{figure}[t]
    \centering
    \begin{subfigure}[t]{0.49\textwidth}
        \centering
        \includegraphics[height=3.5cm]{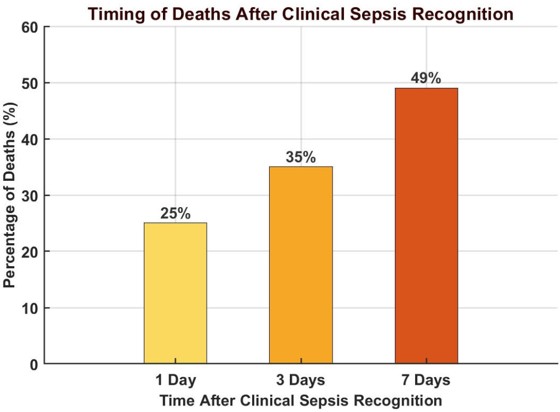}
        \caption{The mortality rate increases rapidly with the time of clinical sepsis diagnosis.}
        \label{fig:mortality-rate}
    \end{subfigure}
    \hfill
    \begin{subfigure}[t]{0.49\textwidth}
        \centering
        \includegraphics[height=3.5cm]{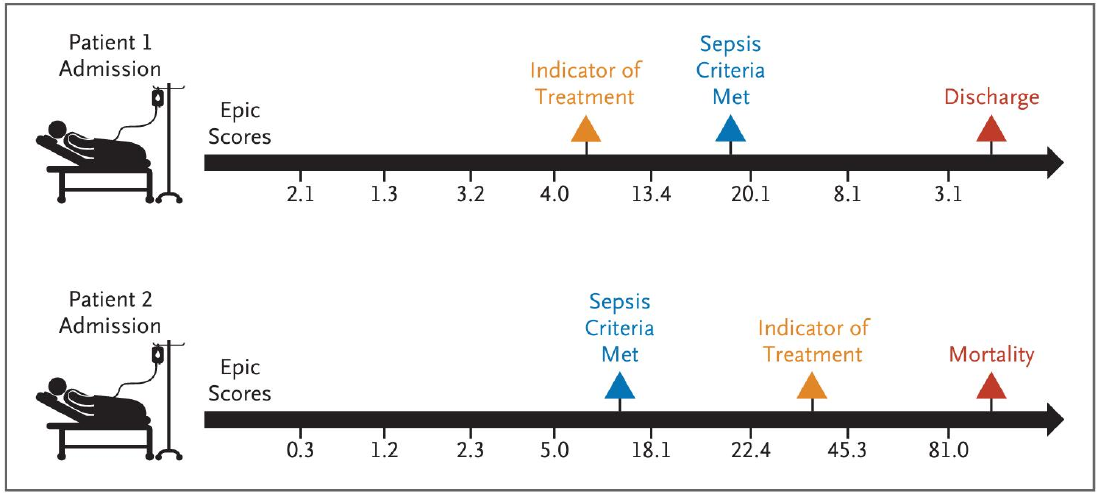}
        \caption{Different conditions of two patients}
        \label{fig:2patients}
    \end{subfigure}
    \caption{Comparison of mortality and early prediction for patients}
    \label{fig:combined}
\end{figure}

Sepsis has two critical characteristics: complex diagnosis and rapid progression. Diagnosing sepsis requires extensive testing, including blood tests, cultures, imaging, and organ function assessments such as the SOFA score~\cite{moreno2023sequential}. These procedures help identify infection sources and evaluate the extent of organ dysfunction. In addition, sepsis progresses quickly—within 12 hours, a patient’s condition can deteriorate severely. According to Masmali et al.~\cite{masmali2024epidemiology}, 25\% of sepsis-related deaths occur within one day of diagnosis, increasing to 49\% by day seven shown by Figure~\ref{fig:mortality-rate}. This highlights the urgent need for early detection and intervention.

Figure~\ref{fig:2patients} illustrates two contrasting cases. Patient 1 received early treatment before meeting clinical criteria and recovered. Patient 2 was treated only after diagnosis, by which point organ failure had likely set in, leading to death. Early prediction is therefore essential for improving patient outcomes.

Therefore, leveraging rapidly advanced AI technologies to develop models that can accurately predict sepsis several hours in advance holds great promise for improving patient outcomes.

\section{Research Gaps}
Despite growing interest in machine learning for sepsis prediction, persistent research gaps hinder its clinical adoption. Studies often differ in data, features, and sepsis definitions, complicating comparisons and reproducibility—only 1 of 28 studies followed standard reporting guidelines \cite{fleuren_machine_2020}. Most models are validated on retrospective North American datasets, limiting generalization to diverse or resource-limited settings \cite{bignami_artificial_2025}. Real-world deployment is rare, with high false-positive rates in EHR-based tools leading to alarm fatigue and clinician mistrust \cite{boussina_impact_2024}.

Crucial challenges remain, including the lack of randomized controlled trials, limited contextual validation in settings like emergency departments, and data quality issues such as missing or inconsistent EHR entries \cite{boussina_impact_2024, bignami_artificial_2025}. Deep learning models face scrutiny over interpretability, with clinicians needing explainable outputs from tools like SHAP or LIME \cite{li_harnessing_2025}. Ethical concerns about biased training data and device inaccuracies raise equity issues, while the long-term impacts of false positives are understudied \cite{smith_detecting_2024}.

Few models undergo robust external validation; widely used tools like the Epic Sepsis Model often perform poorly in new environments \cite{wong_external_2021}. Advancing the field requires standardization, multi-center prospective validation, better data access, and innovations in temporal modeling and ensemble methods to ensure clinical relevance and equity \cite{bomrah_scoping_2024}.

\section{Proposed Method \& Machine Learning Solution}
According to a systematic review by Fleuren et al. (2020)\cite{fleuren_machine_2020}, a wide range of studies have addressed sepsis prediction, including early detection and neonatal mortality. However, these studies often rely on different features and clinical scores, such as the SOFA score, leading to inconsistency. While the SOFA score is useful for assessing organ dysfunction in ICU settings, it may not be optimal for early detection, as it typically identifies sepsis only after significant deterioration has occurred. Early intervention is critical in managing sepsis, and models that depend on overt signs of organ failure may detect the condition too late.
\begin{figure}
    \centering
    \includegraphics[width=0.8\linewidth]{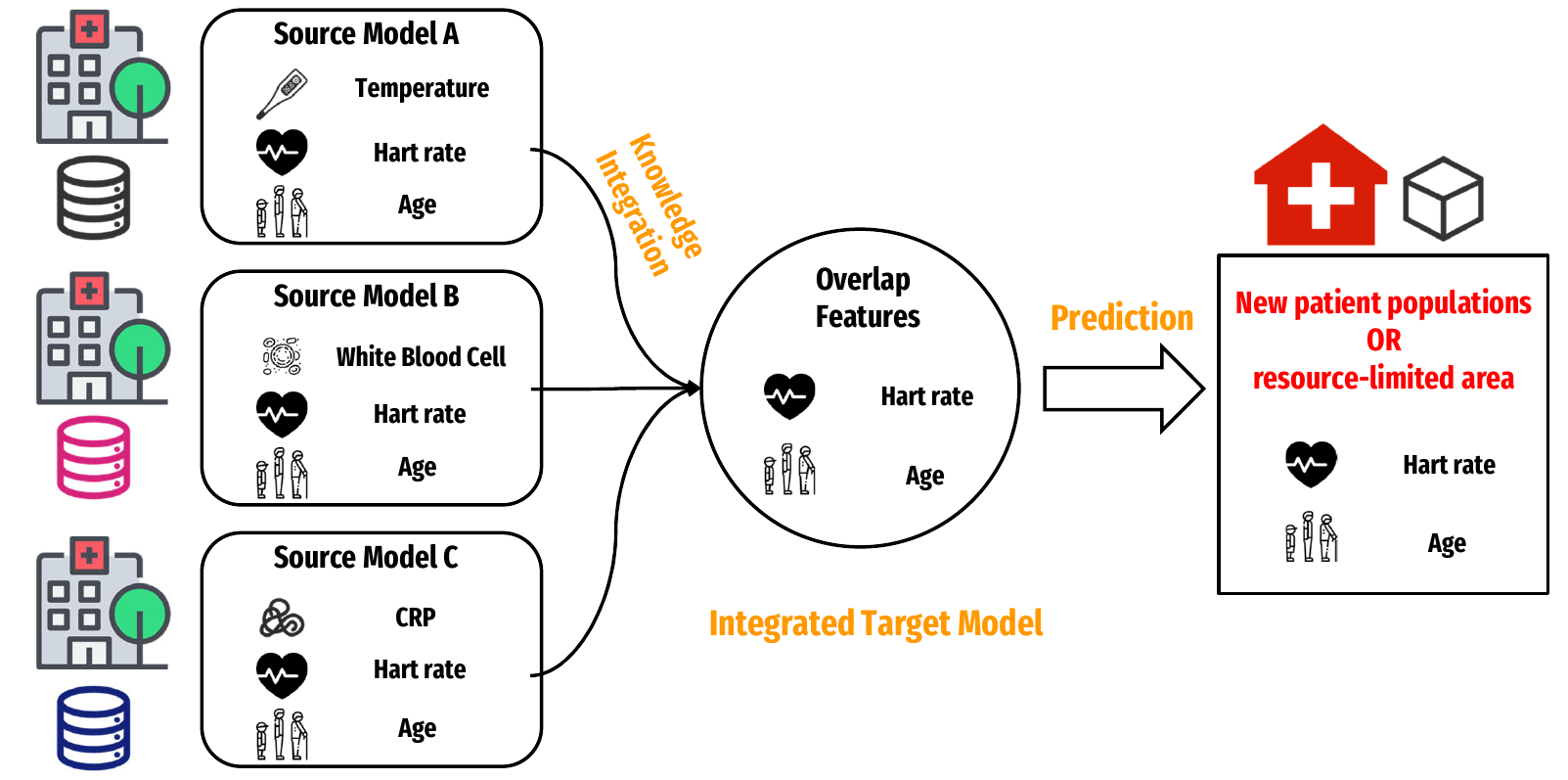}
    \caption{The architecture of our system}
    \label{fig:archi}
\end{figure}
These differences in feature selection and population focus make it difficult to generalize predictive models across datasets and clinical environments. As illustrated in Figure~\ref{fig:archi}, we propose a unified approach that integrates knowledge from multiple source models to improve generalizability. Our method is based on a \textbf{feature-aligned transfer learning (FATL)} framework.

\begin{itemize}
    \item First, we identify biological features that are commonly emphasized across prior models. These shared features act as robust indicators and can help prioritize patient testing, especially in settings with limited medical resources.
    \item Second, we address population bias by leveraging source models trained on diverse cohorts. Instead of relying on a single dataset, we combine model parameters using a weighted scheme to capture a broader representation of patient profiles.
\end{itemize}

By aligning features and aggregating knowledge from diverse sources, FATL enables a more reliable and transferable solution for early sepsis prediction.

\subsection{Feature-Aligned Transfer Learning (FATL)}
\begin{figure}
    \centering
    \includegraphics[width=0.9\linewidth]{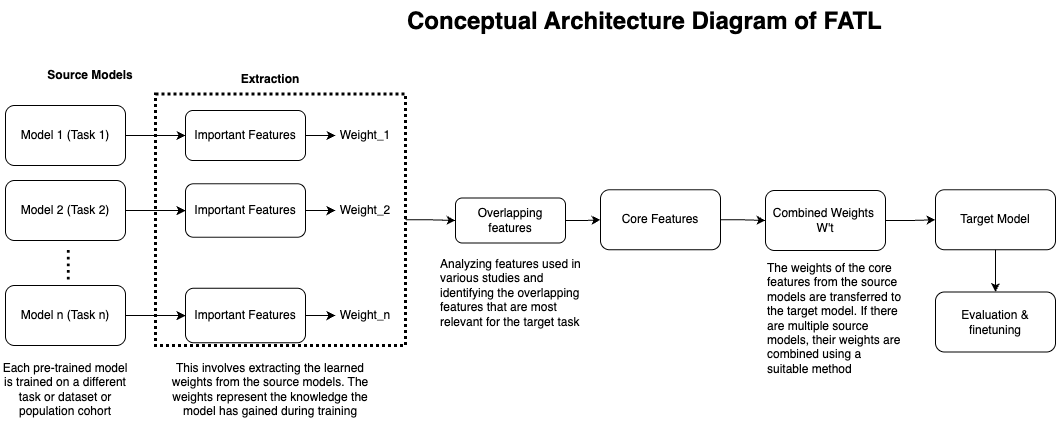}
    \caption{Conceptual Architecture Diagram of FATL}
    \label{fig:FATL}
\end{figure}

FATL (\ref{fig:FATL}) is a machine learning methodology that combines transfer learning with a focus on feature selection. In FATL, the process involves:
\begin{itemize}
    \item Identifying key features that are consistently important across various studies or datasets related to the target prediction task.
    \item Transferring knowledge from source models to a target model, with an emphasis on aligning the transferred knowledge with these key features
\end{itemize}

The goal of FATL is to improve prediction accuracy, generalization, and robustness by focusing on the most relevant information and leveraging knowledge from diverse sources.

\subsubsection{Feature Selection and Harmonization}
This process involves: (1)Review of the literature of the focus area to identify the features considered important in various studies. (2)Analyze the selected studies and identify features that consistently appear as significant predictors. (3)Harmonize the overlapping features to ensure consistency in definition, units or scales across the datasets. (4)Identify and handle data quality

\subsubsection{Weight Transfer}

Weight transfer is a machine learning technique where a model trained on one task is reused for a new task, typically when the new task has less data. It helps the new model start with useful, general features learned from a larger dataset. In sepsis detection, weight transfer allows the combining of models trained on different features, improving sensitivity to early signs and enhancing generalization across diverse patients and clinical settings. Weight transfer for \emph{target model} with various tasks \emph{source models} is defined as:



\begin{equation} \label{eq:1}
    W't = \sum_{i=1}^{n} \alpha_i \cdot W_i 
\end{equation}
 
 $W't$ is the initial weight of the target model after the transfer. 
 
 $\alpha_i$ weight transfer coefficient representing the importance or contribution of the i-th source model. $W_i$ weights from the i-th source model, where (1=1,2...n) i.e weights from the various tasks $n$. 
 
 The weights of each source model $W_i$ are multiplied by a coefficient $\alpha_i$ that determines how much influence that model has on the initial weights of the target model. 
 
 $\alpha_i$ coefficient allows for flexibility in giving more or less importance to different source models based on their relevance to the target task. If $\alpha_i = \frac{1}{n}$ for all (i), it becomes a simple average of all source model weights. 
 
 Equation \ref{eq:1} allows weight transfer from multiple sources and fine-tuning of each source contribution. 

\subsubsection{FATL formula}
Therefore FATL can be defined as:
 
\begin{equation} \label{eq:2}
    W't = \sum_{i=1}^{n} \alpha_i \cdot (W_i \odot F) - \lambda \cdot (b_t - \frac{1}{n} \sum_{i=1}^{n} b_i)
\end{equation}

$F$ is a filter that allows the transfer of weights from source models focused on the most important features. 

$F$ is an array of weights representing the importance of the j-th feature, where (j = 1, 2, ..., m) (total number of features). 

If $F_j$ is close to 1, the corresponding weights are transferred; if it's close to 0, they are suppressed. 

$\alpha_i$ weight transfer coefficient representing the importance or contribution of the i-th source model. 

$\lambda \cdot (b_t - \frac{1}{n} \sum_{i=1}^{n} b_i)$ is the bias regularization term. 

$\lambda$ Bias penalty coefficient, controlling the strength of the penalty. A larger $\lambda$ enforces a stronger similarity between the target model's bias and the source models' average bias. $b_t$ Bias vector of the target model. $b_i$ Bias vector of the i-th source model. 

\subsection{Implementation of FATL for Sepsis Use case}

To begin the process, the PubMed database was searched for publications done between 2020-2025 using the keywords: Machine Learning, Sepsis, Detection/Prediction. Finally, 18 papers were selected for research. From the reviews, the following were observed: (1) Most of the papers didn't report the weights/coefficients of the features. (2) ML models are not well documented and, therefore, not reproducible. Pre-processing done on the features is sparsely documented. (3) When using multiple ML, each model's feature importance isn't disclosed.

\subsection{Preliminary Result}

\begin{figure}
  \centering
  \includegraphics[width=0.8\textwidth]{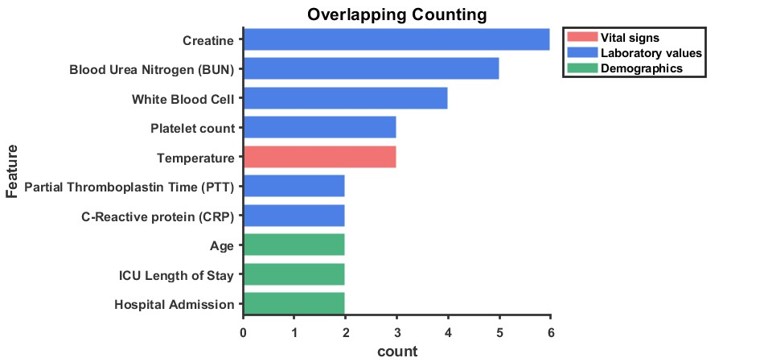}  
    \caption{Overlapping Features}
    \label{fig:Overlapping}
\end{figure}

Based on an analysis of the 18 selected articles, we identified the top 10 most important biological characteristics, as shown in Figure~\ref{fig:Overlapping}. These features fall into three categories: vital signs, laboratory values, and demographic information. They can be considered the key indicators to help in the detection and prediction of sepsis. In resource-limited medical settings, focusing on these features can help prioritize and guide early diagnosis and intervention efforts.

\subsection{Conclusion, Challenges and Limitations}
The FATL (Feature-Aware Transfer Learning) methodology combines  the strengths of weight transfer - leveraging knowledge from diverse data - and feature selection -focusing on the most relevant information - to develop a more accurate and robust model. FATL is applicable to a variety of problems, particularly in healthcare and scenarios where data is scarce or heterogeneous such as medical diagnosis, image recognition and Natural Language Processing.
FATL method can improve accuracy, by focusing on overlapping important features, models can prioritize the most relevant information, potentially leading to higher accuracy. Models can generalize better due to weight transfer from multiple source models and this can lead to faster convergence of model. Subsequently, focusing on clinically relevant and consistently identified features can increase the model's acceptance and utility in clinical practice. On the flip side, FATL relies on weights and known bias to be efficient, therefore, inconsistencies or biases in the source model can affect model performance.
\subsection{GenAI tool usage}
In this report, we used GPT-4o~\cite{OpenAIGPT4Research} to help polish sentences and correct grammatical errors, but we already wrote the content we wanted to express.

\section{Stakeholder Analysis}
If this project advances to clinical testing and implementation, it will require collaboration across multiple domains. Key stakeholders include frontline healthcare providers, who would use the system and provide workflow feedback, and patients and families, who would guide design decisions on transparency and privacy. In Canada, groups like Sepsis Canada involve patient partners in improving detection and care. Hospital leaders, data scientists, IT professionals, and policymakers would support safety, validation, integration, and regulatory compliance. Identifying these stakeholders early lays the groundwork for future implementation.

\bibliographystyle{unsrt}
\bibliography{template}
\end{document}